\documentclass[letterpaper]{article} 
\usepackage{aaai25}  
\usepackage{times}  
\usepackage{helvet}  
\usepackage{courier}  
\usepackage[hyphens]{url}  
\usepackage{graphicx} 
\usepackage{natbib}  
\usepackage{caption} 
\usepackage{algorithm}

\usepackage{amsmath, amssymb, amsfonts, booktabs, enumitem}
\usepackage{algpseudocode}
\usepackage{amsthm}

\usepackage{graphicx, subfig, tikz}
\usepackage[algo2e,ruled,linesnumbered]{algorithm2e}

\usepackage[leftcaption]{sidecap} 

\usepackage{newfloat}
\usepackage{listings}

\newcommand{\X}{\mathbf{X}}
\newcommand{\x}{\mathbf{x}}

\newcommand{\argmax}{\arg \max}

\theoremstyle{plain}
\newtheorem{theorem}{Theorem}

\theoremstyle{definition}
\newtheorem{definition}[theorem]{Definition}

\theoremstyle{remark}

\usepackage{amsmath}

\usepackage{newfloat}
\usepackage{listings}
\DeclareCaptionStyle{ruled}{labelfont=normalfont,labelsep=colon,strut=off} 
\floatstyle{ruled}
\newfloat{listing}{tb}{lst}{}
\floatname{listing}{Listing}
\title{A Unified Framework for Human-Allied Learning of Probabilistic Circuits}
\author {
    Athresh Karanam\equalcontrib,
    Saurabh Mathur\equalcontrib,
    Sahil Sidheekh\equalcontrib,
    Sriraam Natarajan
}
\affiliations {
    The University of Texas at Dallas\\
    \{Athresh.Karanam, SaurabhSanjay.Mathur, Sahil.Sidheekh, Sriraam.Natarajan\}@utdallas.edu
}

\usepackage{bibentry}

\begin{document}

\maketitle

\begin{abstract}
Probabilistic Circuits (PCs) have emerged as an efficient framework for representing and learning complex probability distributions. Nevertheless, the existing body of research on PCs predominantly concentrates on data-driven parameter learning, often neglecting the potential of knowledge-intensive learning, a particular issue in data-scarce/knowledge-rich domains such as healthcare. To bridge this gap, we propose a novel unified framework that can systematically integrate diverse domain knowledge into the parameter learning process of PCs. Experiments on several benchmarks as well as real-world datasets show that our proposed framework can both effectively and efficiently leverage domain knowledge to achieve superior performance compared to purely data-driven learning approaches.
\end{abstract}

%
\begin{links}
    \link{Code}{https://github.com/athresh/unified-constraints-pc/}
\end{links}

\section{Introduction}
Probabilistic Circuits (PCs) have emerged as powerful tools for representing complex probability distributions. Their key advantage is enabling efficient and exact probabilistic inference \cite{ProbCirc20}. Recent work has focused on improving PC expressivity, and allowing them to compete with deep generative models (DGMs) ~\cite{peharz_20_einsum, liu2023scaling, correia2023continuous, sidheekh2023probabilistic}. However, similar to DGMs, this data-driven approach often leads to models that are highly reliant on large datasets, make strong assumptions, and are susceptible to outliers~\cite{ventola2023probabilistic}. 
Purely data-driven PCs often struggle in data-scarce and noisy domains, exhibiting overfitting and reduced generalizability. While traditional techniques like regularization can address overfitting to some extent, they often do so by introducing strong and sometimes arbitrary assumptions about the data distribution~\cite{Domingos1999}. This can limit the model's ability to capture the true underlying relationships between variables.

Incorporating domain knowledge and expert insights into PC learning offers a compelling solution to these problems. Domain experts can provide invaluable insights that go beyond the raw data, grounded in experience and contextual understanding. This knowledge can not only mitigate the risk of overfitting without sacrificing expressive power but also tailor models to specific domain constraints. For instance, fairness and privileged information might be crucial factors in real-world decision-making, requiring models to incorporate such constraints. Knowledge-intensive learning strategies have demonstrably enhanced both discriminative and generative models in various contexts~\cite{KiGBKokel2020,OdomKPLL2015,AltendorfEtAl2005,Campos2008QIBN,YangEtAl2013,Mathur2023KICN}. These models perform more reliably in scenarios with limited or suboptimal data. However, the concept of integrating human expertise and domain knowledge remains {\em relatively unexplored in the context of the general class of PCs}.

This work addresses this problem by introducing a framework for incorporating domain knowledge into PC learning. We first develop a unified framework that allows encoding {\em various types of knowledge} as probabilistic domain constraints. A key aspect of our work is that we can model any constraint {\bf as long as its violation can be measured by a differentiable function}. We then demonstrate how {\bf some} of these constraints, including equality constraints for generalization and privileged information, and inequality constraints for handling class imbalance, monotonicity, and variable interactions, can be seamlessly integrated into PC learning. Our method can be interpreted as {\bf frustratingly easy} as with~\citet{shi2022gradient}, however, it works well and is generalizable, precisely why simpler learning methods should be favored over unnecessarily complex ones~\cite{RudinICML2024,Rudin19}. This approach makes PCs more adaptable to real-world scenarios with limited data and specific modeling requirements. The key tenets of our framework are (1) Expressivity - allowing incorporation of a wide variety of well-known domain knowledge, (2) Effectiveness - consistently improving model performance by leveraging the domain knowledge and (3) Simplicity - allowing domain experts to easily quantify their domain knowledge through equality and inequality constraints.

We make the following key contributions:(1) We propose a unified framework that subsumes several types of domain knowledge and allows encoding them as domain constraints and demonstrate six particular instantiations of these constraints. (2) We propose an augmented objective function that effectively incorporates these domain constraints and can be optimized through generic parameter learning algorithms for PCs. (3) We experimentally validate the added utility of our approach on several benchmark and real-world scenarios where data is limited but knowledge is abundant.

\section{Background}
\textbf{Notations:}
\noindent We use $X$ to denote a random variable and $x$ to denote an assignment of value to $X$. Sets of random variables are denoted as $\X$ and their values as $\x$. We denote the subset of variables in set $k$ as $\X_{k}$ and those not in $k$ as $\X_{-k}$.
We use $\mathcal{M} = (\theta, G)$ to denote a probabilistic circuit having structure $G$ and parameterized by $\theta$. We use $P(\X)$ to denote the joint probability distribution over $\X$. 

\paragraph{Probabilistic Circuits (PCs)} ~\cite{ProbCirc20} are generative models that represent probability distributions in the form of computational graphs
comprising three types of nodes - sums, products, and leaves. Each node in the graph represents a (possibly unnormalized) distribution over a set of variables, known as its scope. The internal nodes in the graph are sums and products. Sum nodes compute a convex sum of the distributions modeled by its children, representing a mixture distribution. 
Product nodes compute a product over the outputs of their children, representing a factorized distribution over their scopes. Leaf nodes encode simple tractable distributions such as a Gaussian. A PC is evaluated bottom up and the output of its root node gives the modeled joint probability density.

The structure of a PC must satisfy certain properties for it to be able to perform exact inference tractably, two of which are \emph{Smoothness} and  \emph{Decomposability}. Smoothness requires that the scope of each child of a sum node be identical. \emph{Decomposability} requires that the scopes of the children of a product node be disjoint.  Formally, a PC $\mathcal{M}$ is a tuple $\langle G, \theta \rangle$ where $G$ is a DAG consisting of sum, product, and leaf nodes. The distributions over a sum node $s$ having scope $S_s$ and product node $p$ with scope $S_p$ are,
$$ P_s(\X_{S_s} = \x_{S_s}) = \sum_{j\in \text{Ch}(s)} \theta_{sj} P_j(\X_{S_j} = \x_{S_j}),$$
$$ P_p(\X_{S_p} = \x_{S_p}) = \prod_{j\in \text{Ch}(p)} P_j(\X_{S_j} = \x_{S_j})$$
where $\text{Ch}(p)$ denotes the children of node $p$. The distribution at a leaf  $l$ having scope $S_l$ is assumed to be tractable and parameterized by $\theta_l.$ Simple distributions such as Bernoulli and Gaussian are commonly used as leaf distributions.
\\
The structure and parameters of PCs can be jointly learned from data. Structure learning algorithms typically use heuristics to recursively learn the PC
\cite{gens2013learnspn,rooshenas14learning,dang20astrudel
,Adel2015LearningTS,peharz13greedy}. Alternatively randomized structures \cite{Mauro17,peharz20a-rat-spn,peharz_20_einsum} that can be easily overparameterized and scaled using GPUs are used, shifting the focus to parameter learning using data-driven techniques, resulting in deep and expressive PCs 
\cite{correia2023continuous,sidheekh2023probabilistic,liu2023scaling,liu2023understanding}.
A detailed review of such parameterizations can be found in \cite{sidheekh2024building}.
However, these advanced PCs often require large amounts of data to learn effectively, limiting their use in scarce and noisy data settings.

\paragraph{Knowledge-based Learning.} The role of domain knowledge as constraints becomes crucial in guiding probabilistic model construction in several domains. 
These constraints concisely encode information about general trends in the domain and serve as effective inductive biases, yielding more useful and more accurate probabilistic models, especially in noisy and sparse domains~\cite{Shavlik1994KiNN,VanDerGaag2004,AltendorfEtAl2005,YangEtAl2013,OdomKPLL2015,KiGBKokel2020,Plajner2020}. 

Prior works have explored incorporating such constraints in PCs in a hard or exact way. Semantic Probabilistic Layers~\cite{ahmed2022semantic} guarantee exact symbolic constraint satisfaction for structured-output prediction but are limited to propositional logic and hard constraints.~\citeauthor{ghandi2024probabilistic} (\citeyear{ghandi2024probabilistic}) adjust the leaves of trained PCs for probabilistic constraint satisfaction but do not consider parameter adaptation during training. In contrast, we aim to learn more accurate PCs from data by exploiting diverse forms of domain knowledge to enhance generalization in noisy or sparse data settings.

While prior works have explored the use of domain knowledge to learn more accurate PCs, they have either been limited to equality constraints~\cite{SPNConstraintsPapantonis2020} or specific sub-classes of PCs~\cite{galindez2020discriminative, Mathur2023KICN}. 
As far as we are aware, diverse forms of domain knowledge have {\em not been used previously to learn the parameters of the general class of PCs}. 
To bridge this gap, we propose a unified framework for representing and learning from diverse forms of domain knowledge including generalization constraints, qualitative influence statements~\cite{AltendorfEtAl2005}, context-specific independence relations~\cite{boutilier}, class imbalance tradeoffs~\cite{Yang2014}, and privileged information~\cite{pasunuri2016learning}.


\section{Learning PCs with Domain Knowledge}
We aim to solve the following problem:
\\
\fbox{
\centering
\begin{minipage}{.95\linewidth}
    \textbf{Given:} Dataset ${\mathcal{D}}$ over random variables ${\X},$ and a specification of domain knowledge $\mathcal{K}$.\\
    \textbf{To Do:} Learn a probabilistic circuit $\mathcal{M}$ that accurately models $P(\X)$.
\end{minipage}
}\\ \\
Mathematically, the above problem can be expressed as the following constrained optimization:
\begin{equation}
\label{eq:opt1}
\begin{aligned}
    \underset{\mathcal{M}}{\argmax}\ \mathcal{L}(\mathcal{M}, \mathcal{D})
    &\text{ s.t. $\mathcal{M}$ does not violate $\mathcal{K}$}
\end{aligned}
\end{equation}
where $\mathcal{L}(\mathcal{M}, \mathcal{D})$ is the log-likelihood of $\mathcal{D}$ with respect to $\mathcal{D}$ and is given by the following equation
\begin{equation}
    \mathcal{L}(\mathcal{M}, \mathcal{D}) = \sum_{\x \in \mathcal{D}} \log P_\mathcal{M}(\X = \x)
\end{equation}
This formulation reduces to standard maximum likelihood estimation when there are no domain constraints ($\mathcal{K} = \emptyset$). Similarly, including a simple constraint on model complexity ($\mathcal{K} = $``$\mathcal{M}$ is not too complex.'') recovers maximum likelihood estimation with a basic regularization term. We first present an encoding scheme to encode commonly used forms of knowledge as constraints on the PC and then we present an algorithm that learns a PC using these constraints.

\subsection{Encoding knowledge as constraints}
We now demonstrate that commonly used forms of knowledge can be represented as either equality or inequality constraints on marginal and conditional queries on the PC. Specifically, we consider generalization, monotonicity, context-specific independence, class imbalance, synergy, and privileged information. We formally define equality and inequality constraints as follows:
\begin{definition}
    An equality constraint $\langle f, g, \mathcal{S} \rangle$ states that $f(\x) = g(\x') \forall \x, \x' \in \mathcal{S},$ where $f(\x)$ and $g(\x)$ are tractable, differentiable functions of the PC and $\mathcal{S} \subseteq \X^2$ is a set of pairs of data points.
\end{definition}

\begin{definition}
    An inequality constraint $\langle f, g, \mathcal{S} \rangle$ states that $f(\x) > g(\x') \forall \x, \x' \in \mathcal{S},$ where $f(\x)$ and $g(\x)$ are tractable, differentiable functions of the PC and $\mathcal{S} \subseteq \X^2$ is a set of pairs of data points.
\end{definition}

We use the case of learning a PC to model the risk of Gestational Diabetes (GD) using data from a multi-center clinical study as a running example. Ideally, such a PC should capture the relationships between various risk factors such as age, race, BMI, family history, genetic predisposition, and exercise level, by modeling their joint distribution. The following forms of domain knowledge can be used for learning:

\begin{enumerate}[wide, labelindent=0pt]
\item \textbf{Generalization constraints (GC).} These constraints capture inherent symmetries in the data distribution, such as exchangeability \cite{ijcai2022exchangeability} and permutation invariance \cite{manzil2017deepsets}.  For instance, subjects from the same center might have similar distributions due to study design.  We can express this as:
\begin{align*}
    \begin{aligned}
        &P(\x) = P(\x'), \quad \forall (\x, \x') \in \mathcal{D}^2\text{ s.t. }sim(\x, \x') 
    \end{aligned}
\end{align*}
where $sim(\x, \x')$ indicates if $\x$ and $\x'$ are similar based on predefined criteria (e.g., belonging to the same center).

\item \textbf{Context-Specific Independence relations (CSI).} These relations are a generalization of conditional independence relationships between variables. For example, BMI might be independent of age only for patients with GD. We can represent this as:
 \begin{align*}
    \begin{aligned}
        P(x_i \mid \x_k) = P(x_i \mid x_j, \x_k), \quad \forall \x \in \text{Dom}(\X) \text{ s.t. } \x_k = c
    \end{aligned}
\end{align*}
where $\x_k = c$ denotes a specific value assignment to variables in $\X_k$, defining the context (e.g., having GD) where independence between $X_i$ and $X_j$ holds. 

\item \textbf{Preference constraints (PF)}. Medical experts might provide a set of probabilistic logical conditions indicative of high GD risk. We denote this set as $R = \{(r_1, p_1), \dots, (r_M,p_M)\}$, where each pair $(r, p)$ consists of a logical condition $r$ over variables $\X$ and its associated probability $p$.  
These constraints can be expressed as
\begin{align*}
    \begin{aligned}
        &P(x_i = 1 \mid \x_{-i}) = p\quad\forall \x \models r,\ \forall (r,p) \in R
    \end{aligned}
\end{align*}

  \item \textbf{Class-imbalance tradeoff (CT)} In GD screening, minimizing false negatives is crucial since patients predicted as low risk might not undergo further clinical testing. We can express such false negative constraints as
    \begin{align*}
        \begin{aligned}
            t_0 > P(X_i = 0 \mid \x_{-i}), \quad\forall \x \in \mathcal{D} \text{ s.t. } x_i = 1
        \end{aligned}
    \end{align*}
    where $t_0$ is the threshold such that $P(X_i = 0 \mid x_{-i}) > t_0$ is predicted as negative. These constraints ensure that the model is cautious in predicting a negative outcome, thus reducing the risk of false negatives. Constraints to minimize false positives can be formulated similarly.
    \item \textbf{Monotonic Influence Statements (MISs)} The statements express positive or negative monotonic relationships between pairs of variables. For example, the risk of GD might increase with an increase in age.
    We can represent positive monotonic influence of $X_j$ on $X_i$ (${X_j}_\prec^{M+} X_i$) as:
    \begin{align*}
        \begin{aligned}
            P(X_i \leq x_i \mid x_{j}) > P(X_i \leq x'_i \mid x'_{j}) \\ \quad\forall \x, \x' \text{ s.t. } x_i = x'_i,\ x'_j > x_j
        \end{aligned}
    \end{align*}
   (6)  Synergistic influence statements can be encoded similar to monotonicities while (7) Privileged information can be encoded similar to CSIs. We defer the details about both of these constraints to the supplementary material.
    
\end{enumerate}

\subsection{Defining the penalty function}
Having demonstrated that {\bf seven commonly used forms of knowledge} can be represented as equality or inequality constraints on marginal and conditional queries on the PC, we now present a learning algorithm that uses these constraints to learn PCs. To formalize the constrained optimization problem, we define the notion of a \textit{penalty function} that quantifies the severity of domain knowledge violation in the distribution induced by the PC.
\begin{definition}
    A function $\zeta: M \mapsto \mathbb{R}^+_0$ is a penalty function corresponding to domain knowledge $\mathcal{K}$ if it maps a PC $\mathcal{M} \in M$ to a non-negative real number $\zeta(\mathcal{M})$ that quantifies the extent of the violation of knowledge $\mathcal{K}$ such that 
    \begin{enumerate}
    \item $\zeta(\mathcal{M}) = 0$ $\iff$ $\mathcal{M}$ does not violate $\mathcal{K}.$
    \item $\zeta(\mathcal{M}) < \zeta(\mathcal{M}')$ $\iff$ $\mathcal{M}$ violates $\mathcal{K}$ to a lesser extent than $\mathcal{M}'.$
    \item The total violation in a model $\mathcal{M}$ due to two penalty functions $\zeta$ and $\zeta'$ is $\zeta(\mathcal{M}) + \zeta'(\mathcal{M})$
\end{enumerate}

\end{definition}

Using this notation, we can rewrite the optimization problem in equation \eqref{eq:opt1} as
\begin{equation}
\label{eq:opt2}
\begin{aligned}
    \underset{\mathcal{M}}{\argmax}\ \mathcal{L}(\mathcal{M}, \mathcal{D})
    &\text{ s.t. $\zeta(\mathcal{M}) = 0$}
\end{aligned}
\end{equation}
where $\zeta$ measures the model's deviation from domain knowledge $\mathcal{K}.$ For computational reasons, we focus on learning the model parameters for a PC with a given structure ($\mathcal{G}$). The optimization problem becomes finding PC parameters ($\theta$) that maximize the likelihood while adhering to the domain knowledge:
\begin{equation}
\label{eq:opt3}
\begin{aligned}
    \underset{\theta}{\argmax}\ \mathcal{L}(\langle \mathcal{G}, \theta \rangle, \mathcal{D})
    &\text{ s.t. $\zeta(\langle \mathcal{G}, \theta \rangle) = 0$}
\end{aligned}
\end{equation}
However, since domain knowledge elicited from experts might not be perfectly consistent or fully compatible with the given PC structure, equation \eqref{eq:opt2} might not have any feasible solution. We address this by using the penalty method to find a solution closest to the feasible region~\cite{Luenberger2016,Mathur2023KICN}. Algorithm 1 presents the parameter learning procedure, which solves a series of optimizations of the form
\begin{equation}
\label{eq:opt4}
\begin{aligned}
    \theta_t^* = \underset{\theta}{\argmax}\ \mathcal{L}(\langle \mathcal{G}, \theta \rangle, \mathcal{D}) - \lambda_t \zeta(\langle \mathcal{G}, \theta \rangle) 
\end{aligned}
\end{equation}
where $\theta_t^*$ is the solution of the $t$th optimization and $\lambda_t$ is the penalty weight in the $t$th optimization. The initial value of the penalty weight, $\lambda_0$ is set to $0$ (corresponding to the maximum likelihood solution), and the value of $\lambda_t$ for each subsequent optimization problem is increased using the penalty control parameter $\gamma$ and is set to $\gamma^{t-1}.$ Each optimization where $t > 0$ is initialized with the solution from the previous problem $\theta_{t-1}$. Note that while the objective function appears simple (as a function of likelihood with penalty), this has been widely used for structure learning in graphical models~\cite{schwarz1978bic}, and more recently, even in domain adaptation. Our empirical evaluation is yet another proof that one does not necessarily need complex objective functions if the simpler ones are both effective and by definition, efficient.
\begin{algorithm2e}[t]
  \SetAlgoLined
  \caption{Learn PC Parameters with Knowledge}
  \label{alg:FitParametersWithKnowledge}
  \SetKwInOut{Input}{input}\SetKwInOut{Output}{output}
  \Input{Structure of PC $G$,
  Data $\mathcal{D}$, 
  Penalty violation function $\zeta: M \mapsto \mathbb{R}^+_0$, \\
  Maximum number of tries $t_\text{max}$,
  Penalty weight control $\gamma$,} 
  \Output{Parameters for PC  $\theta$}
  \textbf{initialize:} $\theta = \underset{\theta}{\arg \max}\ \mathcal{L}(\theta; G, \mathcal{D}), t = 1$  \algorithmiccomment{start with maximum likelihood solution}

  $\lambda_1 = 1$
    
  \While{${\zeta(G, \theta) \neq 0}$ and $t \leq t_\text{max}$}{ \algorithmiccomment{while constraints are not satisfied} 
  
    $\theta = \underset{\theta}{\arg \max}\ \mathcal{L}(\theta; G, \mathcal{D}) - \lambda_t \zeta(G, \theta)  $

    $\lambda_{t+1} = \lambda_{t} \times \gamma$ \algorithmiccomment{increase penalty weight} 
    
    $t = t + 1$

  }
  
  \textbf{return} $\theta$
\end{algorithm2e}

\setcounter{theorem}{0}
\begin{theorem}
    If the PC $\mathcal{M}$ is smooth, decomposable, and deterministic, and $\zeta$ is a differentiable, concave function of $\theta,$ then algorithm 1 is guaranteed to converge to the optimal feasible solution of equation \eqref{eq:opt3}, if one exists, as $t_\text{max} \rightarrow{} \infty.$
\end{theorem}
\begin{proof}
    The likelihood of deterministic PCs is a concave function of the parameters~\cite{ProbCirc20}. The penalty method is guaranteed to converge to an optimal solution of the constrained optimization problem if one exists, the objective function is concave and the penalty function is differentiable~\cite{Luenberger2016}. 
\end{proof}

We define the penalty function for equality constraints as $\zeta(G, \theta) = \sum_{(\x,\x')\in\mathcal{S}}|\delta(\x,\x')|$ and for inequality constraints as  $\zeta(G, \theta) = \sum_{(\x,\x')\in\mathcal{S}}\text{max}\{0,\delta(\x,\x') + \epsilon\}^2.$ Here, $\delta(\x,\x') = g(\x)-f(\x)$ and $\epsilon > 0$ is a margin parameter used to control the minimum extent of the inequality. In practice, we might approximate the exact constraint using a subset of $\mathcal{S'}\subseteq\mathcal{S}$ of size $\gamma_\text{size}.$ In the case of multiple pieces of knowledge, we solve the constrained optimization stage-wise. We add constraints corresponding to each piece of knowledge one by one, using the solution at each stage as initialization for the next. We use algorithm \ref{alg:FitParametersWithKnowledge} to find the solution to the sub-problem at each stage by defining the penalty function $\zeta$ as the sum of the penalty functions corresponding to the current set of constraints.

\section{Experimental Evaluation}
Our framework is generic and agnostic to the type of PC used. Thus, to empirically validate the effectiveness of incorporating domain constraints we consider two different instantiations of deep PCs - (i) \textit{RatSPN} \cite{peharz20a-rat-spn} and (ii) \textit{EinsumNet} \cite{peharz_20_einsum}. We implement both models with and without domain constraints for a comparative analysis. We design experiments over various datasets and scenarios to answer the following questions empirically:
\begin{itemize}[wide, labelindent=0pt]
    \item[({\bf Q1})] Can domain knowledge be incorporated faithfully into the parameter learning of PCs?
    \item[({\bf Q2})] Does incorporating knowledge improve the generalization performance of PCs?
    \item[({\bf Q3})] Can the proposed framework exploit domain knowledge in heterogenous forms to learn more accurate PCs?
    \item[({\bf Q4})] How sensitive is the approach to the hyperparameters governing the size of domain sets and penalty weight? Further, is it robust to noisy or redundant advice?
    \item[({\bf Q5})] Does the proposed method learn more accurate models on real-world data?
\end{itemize}

\vspace{-0.5em}
\paragraph{Experimental Setup.} To answer these questions, we compare our knowledge-intensive learning method to purely data-driven approaches. We defer further implementation details to the supplementary material. We used three types of data sets for our experiments: synthetic, benchmark, and real-world clinical data.

\textbf{Synthetic data sets.} We used two types of synthetic data sets: four Bayesian Network (BN)-based datasets and a manifold-based data set. We derived data sets from 4 BNs: Earthquake~\cite{EarthquakeKorb2010}, Asia~\cite{AsiaLauritzen1988},  Sachs~\cite{Sachs2005}, and Survey~\cite{scutari2021bayesian}. For each BN, we constructed a data set by sampling 100 data points. To simulate perfect domain knowledge, we then extracted two conditional independence relations from each BN. These relations were then translated into CSI constraints. 

\begin{figure}[t]
    \centering
    \begin{tabular}{ccc}

    \hspace{-2.0em}
    \subfloat[Train]{ \includegraphics[width=0.24\linewidth]{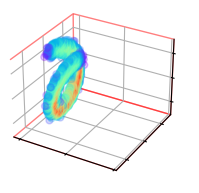}}
    \hspace{-1.0em}
    
    \subfloat[Test]{\includegraphics[width=0.24\linewidth]{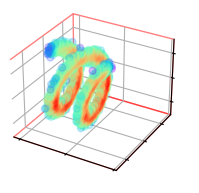}}
    
    \subfloat[\textit{EinsumNet}]{\includegraphics[width=0.24\linewidth]{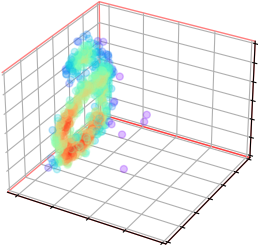}}
    \hspace{0.5pt}
    \subfloat[\textit{EinsumNet+GC}]{\includegraphics[width=0.3\linewidth]{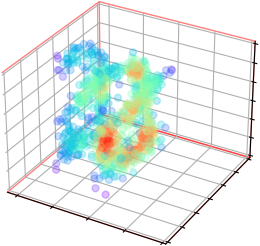}}
   
    \end{tabular}
    \caption{\small \textbf{3D Helix dataset}: Visualization of the 3D Helix dataset. The train dataset (a) consists of a single helix of length $2\pi$ with Gaussian noise added to it. The test  dataset (b) consists of a helix of length $4\pi$ with Gaussian noise added to it. (c) and (d) visualizes 1000 samples generated from \textit{EinsumNet} with and without incorporating the Generalization Constraint (\textit{GC}), respectively.}
    \label{fig:3D_helix}
\end{figure}

We used a 3D Helix Manifold to construct the second type of synthetic data set (Figure~\ref{fig:3D_helix})~\cite{vqflows-sidheekh22a}. This dataset presents difficulties due to its intricate spiral structure. However, its inherent symmetry can be exploited to define a generalization constraint (GC). We encoded the helical symmetry by specifying that points separated by $2\pi$  along the x-axis and lying on a helix manifold of unit radius have similar density. We use 100 pairs of datapoints of the form $((x,sin(x),cos(x)), (x+2\pi,sin(x+2\pi), cos(x+2\pi))).$

\textbf{Benchmark data sets.} We used two kinds of benchmark data sets -- four UCI benchmark data sets and two point cloud-based data sets. We used 4 data sets from the UCI Machine Learning repository, namely, breast-cancer, diabetes, thyroid, and heart-disease. We used the monotonic influence statements that were used in prior work by~\citeauthor{YangEtAl2013} (\citeyear{YangEtAl2013}) as domain knowledge for these data sets.

We also used the MNIST~\cite{mnist} and fashion-MNIST image datasets to construct point cloud representations following ~\cite{zhang2019deep}. We transformed each image into a set-based representation by sampling the locations of a small set of foreground pixels. The resulting 2D point cloud representation inherently exhibits permutation invariance – a symmetry difficult to model even for advanced generative models ~\cite{li2018layoutgan,kim2021setvae} such as Generative Adversarial Networks and Variational Autoencoders. We refer to these datasets as full set data (e.g., Set-MNIST-Full). Additionally, to simulate a data-scarce scenario, we further divided the MNIST dataset into two subsets: (i) Set-MNIST-Even comprising only even digits and (ii) Set-MNIST-Odd comprising only odd digits. We encoded permutation invariance as a GC by specifying the domain set criteria as $sim(\x,\x')= \mathbb{I}[\x'=\pi(\x)]$, where $\pi(\x)$ represents a permutation of $\x$. In practice, we defined the GC domain set by taking $\gamma_{size}=2$ permutations for each sample in the dataset.

\textbf{Real-World Clinical Data.} To evaluate the performance of our framework on real-world data, we used data from the Nulliparous Pregnancy Outcomes Study: Monitoring Mothers-to-Be (nuMoM2b,~\cite{numom2b}). We considered two subsets of the data, numom2b-a focusing on a single Adverse Pregnancy Outcome (APO) namely Gestational diabetes (GD), and numom2b-b focusing on three different APOs namely New Hypertension (NewHTN), Pre-eclampsia (PreEc), and Pre-term Birth (PTB). 
\\
The first data set consists of 3,657 subjects of white, European ancestry that have data for GD and its $7$ risk factors: Age, BMI at the start of pregnancy, presence of Polycystic Ovary Syndrome (PCOS), physical activity (METs), presence of High Blood Pressure (HiBP), family history of diabetes (Hist), and a polygenic risk score (PRS) indicating genetic predisposition to diabetes. All risk factors except physical activity (METs) are known to positively monotonically influence the risk of GDM, while physical activity is known to have a negative monotonic influence.
The second data set consists of 9,368 subjects of diverse racial and ethnic backgrounds that have data for three APOs: NewHTN, PreEc, PTB, and four risk factors: BMI, Age, Hist, and HiBP.
\subsection{Results}
\begin{enumerate}[wide, labelindent=0pt]
    \item[({\bf Q1})]\textbf{Faithful Incorporation of Domain Constraints.} 
    
\begin{table}[t]
\resizebox{\columnwidth}{!}{
\begin{tabular}{@{}clll@{}}
    \toprule
    &  &\multicolumn{1}{c}{\textit{RatSPN}} & \multicolumn{1}{c}{\textit{RatSPN+Constraint}} \\ \midrule

    BN & asia          & $-483.3   \pm 4.1$  & $\mathbf{-313.2  \pm 3.9}$ \\
       &sachs          & $-1097.5 \pm 8.8$  & $\mathbf{-861.2 \pm 8.7}$ \\
       &survey & $-611.7 \pm  7.2$ & $\mathbf{-476.6 \pm 6.6}$ \\
       &earthquake     & $-272.0 \pm 2.4$ & $\mathbf{-121.8  \pm 2.1}$ \\
    \midrule
    UCI&breast-cancer  & $-2110.8 \pm 15.6$ & $\mathbf{-1271.5 \pm 14.6}$   \\
       &diabetes       & $-7010.3 \pm 31.0$ & $\mathbf{-5070.3 \pm 481.8}$\\
       &thyroid        & $ -351.5  \pm 6.1$  & $\mathbf{-200.5  \pm 23.2}$\\
       &heart-disease  & $-931.7  \pm 15.0$ & $\mathbf{-739.8  \pm 7.2}$ \\ 
    \midrule
    RW &numom2b-a        & $-14573.9 \pm 69.9$ & $\mathbf{-7288.2  \pm  1.6}$ \\ \bottomrule
    \end{tabular}
}
    \caption{\small\textbf{Quantitative Evaluation:} Mean test log-likelihood of \textit{RatSPN} trained with and without constraints on the Bayesian Network (BN), UCI Benchmark and Real-World (RW) datasets, $\pm$ standard deviation across $3$ independent trials.}%
\label{tab:bn-uci}
\end{table}

    Table~\ref{tab:bn-uci} (rows 1-4) presents the test log-likelihood scores for RatSPN models trained on the BN-based data sets with and without domain knowledge encoded as CSIs. Notably, the degree of constraint violation for RatSPN models incorporating domain constraints remained consistently below $0.0001$ across all datasets. This confirms our framework's ability to faithfully integrate valid domain knowledge. Moreover, RatSPN models trained with domain constraints performed better than those trained solely on data.

    \item[({\bf Q2})]\textbf{Improvement in Generalization Performance.}

\begin{table*}[t]
\centering
\begin{tabular}{@{}llllll@{}}
\toprule
             & \multicolumn{1}{c}{Helix} & \multicolumn{1}{c}{Set-MNIST-Even} & \multicolumn{1}{c}{Set-MNIST-Odd} & \multicolumn{1}{c}{Set-MNIST-Full} & Set-Fashion-MNIST      \\ \midrule
\textit{RatSPN}       & $-7.1 \pm 1.8$           & $-657.7 \pm 15.9$              & $-682.3 \pm 16.7$             & $-599.3 \pm 3.2$               & $-1495.0 \pm 3.5$  \\
\textit{RatSPN+GC}    & $\mathbf{-3.6 \pm 0.1}$           & $\mathbf{-562.2 \pm 0.4}$               & $\mathbf{-555.1 \pm 0.4}$              & $\mathbf{-566.1 \pm 0.1}$               & $\mathbf{-1236.9 \pm 0.4}$  \\ \midrule
\textit{EinsumNet}    & $-5.1 \pm 0.2$           & $-666.2 \pm 13.7$              & $-683.7 \pm 19.9$             & $-600.8 \pm 4.8$               & $-1413.9 \pm 20.8$ \\
\textit{EinsumNet+GC} & $\mathbf{-2.7 \pm 0.1}$           & $\mathbf{-561.9 \pm 0.2}$               & $\mathbf{-555.1  \pm 0.1}$             & $\mathbf{-566.7 \pm 0.3}$               & $\mathbf{-1235.8 \pm 0.6}$   \\ \bottomrule
\end{tabular}
\caption{\small\textbf{Quantitative Evaluation}: Mean test log-likelihood of \textit{RatSPN} and \textit{EinsumNet} trained with and without the Generalization Constraint (\textit{GC}) on the Helix and Set-MNIST datasets, $\pm$ standard deviation across $3$ independent trials.}
\label{tab:generalization-performance}
\end{table*}

\begin{table}[ht]
\centering
\resizebox{\columnwidth}{!}{
\begin{tabular}{llll}
\toprule
           & RatSPN & +CSI & +CSI+MIS \\
\midrule
earthquake & $-272.0 \pm 2.4$ & $-137.7 \pm 4.7$ & $\mathbf{-106.1 \pm 1.1}$ \\
survey &  $-611.7 \pm  7.2$ &  $-523.5 \pm 4.3$ &  $\mathbf{-470.9 \pm 6.6}$\\
asia       & $-483.3 \pm 4.1$ &  $-320.5 \pm 9.9$ & $\mathbf{-284.7 \pm 6.4}$ \\
\midrule
numom2b-b  & $-18281.2 \pm  218.8$ & $-15122.9 \pm 201.7$ & $\mathbf{-14758.1 \pm 60.3}$         \\
\bottomrule

\end{tabular}
}
\caption{\small The test log-likelihoods for the RatSPN learned from data, with one form of knowledge and with two forms of knowledge averaged over 3 independent trials}
\label{tab:multiple_constraints}
\end{table}

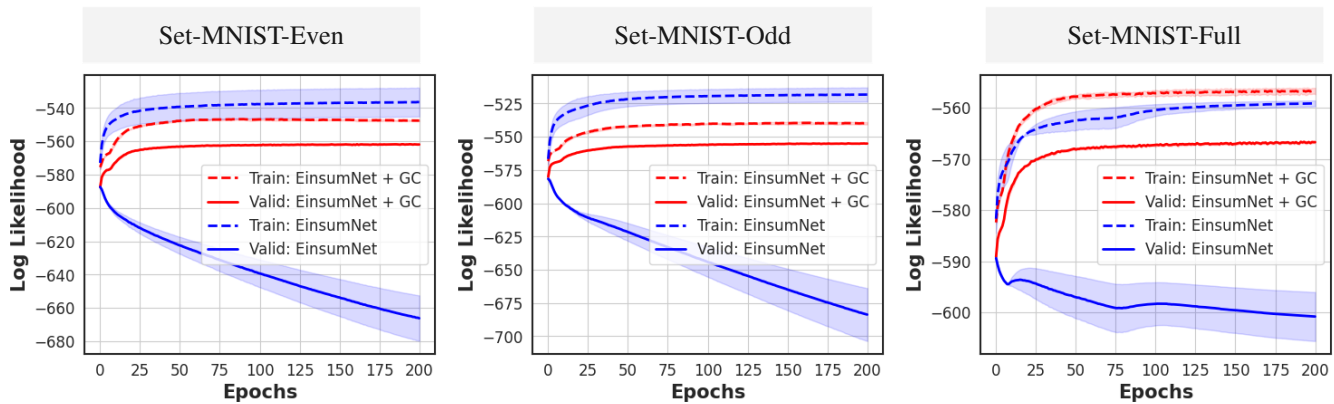
\begin{figure*}
    \centering
    \begin{tikzpicture} 
    \node[rectangle,draw=white!90,fill=white,opacity=1,minimum width=2.0cm,minimum height=0.75cm] at (0,0) {};
    \node[rectangle,draw=white!90,fill=gray!10,minimum width=3cm,minimum height=0.75cm] at (1.25,0) {Set-MNIST-Even};
    \node[rectangle,draw=white!90,fill=gray!10,minimum width=3cm,minimum height=0.75cm] at (7.1,0) {Set-MNIST-Odd};
    \node[rectangle,draw=white!90,fill=gray!10,minimum width=3cm,minimum height=0.75cm] at (12.75,0) {Set-MNIST-Full};
    \end{tikzpicture}
    \includegraphics[width=0.32\linewidth]{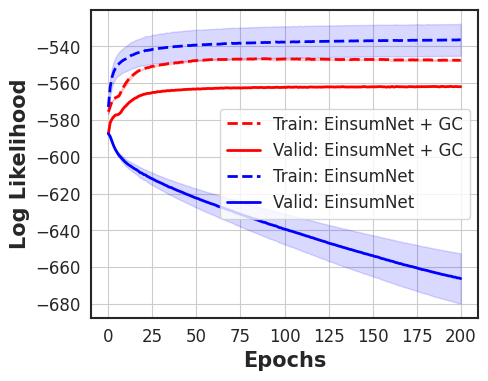}\
    \includegraphics[width=0.32\linewidth]{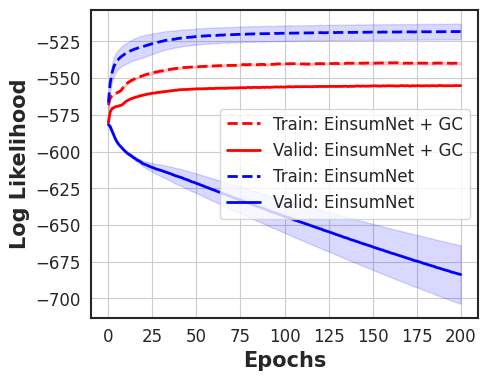}
    \includegraphics[width=0.32\linewidth]{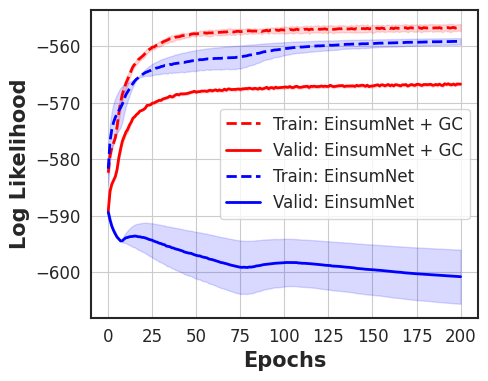}
    \caption{\small
    \textbf{Learning Curves:} Mean train and validation log-likelihoods of \textit{EinsumNet} trained \textcolor{red}{with} (in red) and \textcolor{blue}{without} (in blue) incorporating Generalization Constraint (\textit{GC}) on the three Set-MNIST datasets, across training epochs. 
    The shaded regions denote the standard deviation across $3$ independent trials.}
    \label{fig:set-mnist-learning-curves-einet}
\end{figure*}

    We studied the effect on generalization performance due to two kinds of domain knowledge -- generalization constraints (GC) and monotonic influence statements.
     We used the 3D Helix Manifold dataset and the point cloud data sets to evaluate the effect of GC. Table \ref{tab:generalization-performance} shows the mean test log-likelihoods of the PCs trained on the 3D Helix Manifold data with and without domain knowledge encoded as GC.  The baselines, EinsumNet and RatSPN, simply append the GC data points to the training data. In contrast, our proposed approach (EinsumNet+GC and RatSPN+GC) treats the GC as a constraint during training.   
     We observed significant performance improvements for both models when incorporating the GC. This suggests that the GC enables the models to exploit the inherent symmetries within the data, allowing them to generalize to unseen regions with similar structure. 
     Notably, this improvement required only a small number of samples from the constraint's domain set.
     Additionally, including these data points directly in the training data without structured GC integration did not lead to better generalization (Table~\ref{tab:generalization-performance}). Figure~\ref{fig:3D_helix} visually confirms this, as samples generated by EinsumNet+GC more closely resemble the test data distribution compared to those from EinsumNet.
     
Similar results were observed for set-based image datasets (MNIST and fashion-MNIST) (Table~\ref{tab:generalization-performance}). Incorporating GC significantly improved performance for both models. This is further supported by the quality of samples generated with and without GC (Figure~\ref{fig:qualitative-set-mnist-dataset}). The impact of GC on model training is also evident in the learning curves for Set-MNIST datasets (Figure~\ref{fig:set-mnist-learning-curves-einet}). In the absence of GC (particularly for Set-MNIST-Even and Set-MNIST-Odd with fewer data points), the model overfits as shown by the increasing training log-likelihood and decreasing validation log-likelihood. Conversely, incorporating GC enables the model to leverage symmetries and achieve better generalization. Similar visualizations for other datasets are provided in the supplementary material.
    \begin{figure}[h]
    
     \centering
    \begin{tabular}{cc}
    \subfloat[\textit{EinsumNet}]{ \includegraphics[width=0.45\linewidth]{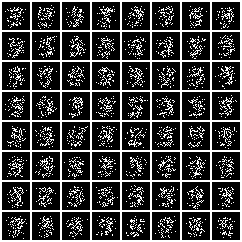}}\hspace{2.5pt}
    \subfloat[\textit{EinsumNet+GC}]{\includegraphics[width=0.45\linewidth]{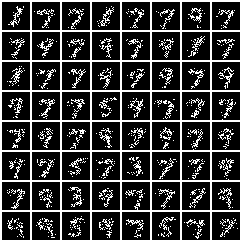}}\hspace{2.5pt}
    \end{tabular}
    \caption{ \small \textbf{Qualitative Evaluation}: Visualization of randomly sampled datapoints generated by \textit{EinsumNet} trained with and without the Generalization Constraint (\textit{GC}) on Set-MNIST-Odd.}
    \label{fig:qualitative-set-mnist-dataset}
    \end{figure}
    
    To evaluate the effect of monotonic influence statements, we used four UCI benchmark datasets. Table~\ref{tab:bn-uci} shows the improvement in the test log-likelihood of RatSPN trained on UCI data sets when incorporating domain constraints.

    \item[({\bf Q3})]\textbf{Incorporation of different types of knowledge}. We studied the effectiveness of our unified framework in incorporating domain knowledge presented in heterogeneous forms. We considered knowledge in the form of CSI and MIS in 3 BN-based domains and 1 clinical domain. Table \ref{tab:multiple_constraints} presents the test log-likelihood scores of RatSPN learned from each of the 4 domains without knowledge (RatSPN), with a single CI encoded as CSIs (+CSI), and with the CSIs and a single MIS (+CSI+MIS). The models using both forms of knowledge perform better than models using CSIs.  
    
    \item[({\bf Q4})]\textbf{Sensitivity and Robustness.}
    Our framework relies on two key hyperparameters: the size of the domain sets used for constraint evaluation (denoted by $\gamma_\text{size}$), and the penalty weight ($\lambda$) that balances constraint satisfaction with data-driven learning. In real-world applications, domain knowledge might be imperfect or noisy. We conducted ablation studies to assess the framework's sensitivity and robustness to these factors.
    $\gamma_\text{size}$ is defined as the ratio of the domain set size for a constraint to the overall dataset size. To simulate noisy domain knowledge, we replaced a fraction of the domain set ($\gamma_\text{noise}$) with randomly sampled data points.

    We investigated the impact of these hyperparameters on an EinsumNet model trained on the Set-MNIST-Even dataset with GC for 100 epochs. Varying $\gamma_\text{size}$ (Figure~\ref{fig:ablation-set-mnist}a) showed that increasing the domain set size generally improves generalization performance, but this effect plateaus after a certain point (around $\gamma_\text{size}\approx 2$).
    Interestingly, varying $\gamma_\text{noise}$ (Figure~\ref{fig:ablation-set-mnist}b) revealed that the model's performance remained relatively stable even with up to 40\% noise in the constraints. This suggests the framework's robustness, as data compensates for some level of noise in the knowledge.
    
    Finally, varying the penalty weight $\lambda$ (Figure~\ref{fig:ablation-set-mnist}c) demonstrated that a very small $\lambda$ leads to underutilization of the constraint, resulting in poor performance. Conversely, a very large $\lambda$ overpowers the maximum likelihood training, disrupting the balance. The optimal $\lambda$ appears to be around 1, striking a balance between data and constraints. 
    \item[({\bf Q5})]\textbf{Performance on Real-World Data.} We evaluated the effectiveness of our framework on real-world data using the numom2b data sets.  Tables \ref{tab:bn-uci} and \ref{tab:multiple_constraints} compare the test log-likelihood of \textit{RatSPN} learned from the numom2b-a and numom2b-b data set with and without the knowledge in the form of MISs and CSIs.  RatSPNs learned with knowledge achieved significantly higher test log-likelihood scores. This demonstrates the efficacy of our framework in exploiting domain knowledge to learn PCs from real-world datasets.
    
    \begin{figure}[t]
        \centering
        \begin{tabular}{ccc}
        \subfloat[Varying $\gamma_{size}$]{ \hspace{-15pt}
        \includegraphics[width=0.33\linewidth]{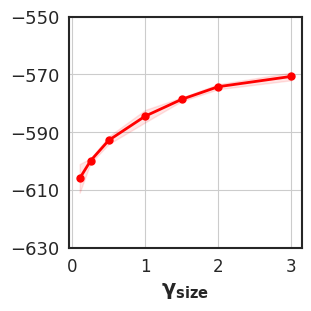}}
        \subfloat[Varying $\gamma_{noise}$]{\hspace{-2pt}\includegraphics[width=0.33\linewidth]{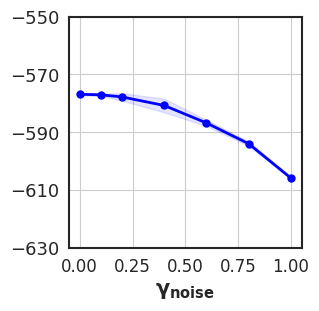}}
        \subfloat[Varying $\lambda$]{\hspace{-2pt}\includegraphics[width=0.33\linewidth]{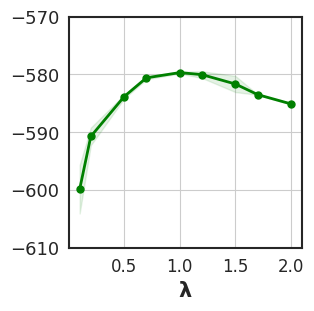}}
        \end{tabular}
        \caption{  \textbf{Ablation Study:} Mean test log-likelihood of a \textit{RatSPN} trained on the Set-MNIST-Even dataset, across varying (a) size of domain sets and (b) degrees of noise. Shaded regions represent the SD across $3$ trials.} 
        \label{fig:ablation-set-mnist}
    \end{figure}

\end{enumerate}

\section{Discussion}
We presented a general approach for incorporating domain knowledge as differentiable constraints into learning probabilistic models.  It subsumes and extends several prior works on learning generative models (e.g., \cite{AltendorfEtAl2005,SPNConstraintsPapantonis2020,Mathur2023KICN}) and (probabilistic) discriminative models  \cite{KiGBKokel2020}.  Prior work by \citeauthor{SPNConstraintsPapantonis2020} (\citeyear{SPNConstraintsPapantonis2020}) focused specifically on knowledge encoded as equality constraints, which is a special case within our broader framework. We also generalize ~\citeauthor{Mathur2023KICN}(\citeyear{Mathur2023KICN})'s approach of using monotonicity constraints to learn cutset networks (CNs) both in the forms of knowledge that can be captured and the model class,  as CNs can be represented as deterministic PCs.
 
\citeauthor{AltendorfEtAl2005} (\citeyear{AltendorfEtAl2005}) used \textit{ceteris paribus}\footnote{Other parents of the influenced variable $X_i$ are kept constant} monotonicity to learn the parameters of Bayesian Networks (BNs). Similar to CNs, BNs can be compiled into selective PCs ~\cite{peharz2014learning}, making our framework applicable here as well. The KiGB algorithm~\cite{KiGBKokel2020} used soft monotonicity constraints for learning decision trees and decision trees can be converted into conditional SPNs \cite{Shao20CSPN}. These methods can be seen as special cases of our framework. 

Our framework is also versatile enough to incorporate more complex forms of knowledge such as \cite{OdomKPLL2015,Xu18SemanticLoss}, enhancing its applicability compared to previously discussed methods. \citeauthor{OdomKPLL2015} (\citeyear{OdomKPLL2015}) used relational preference information to learn gradient-boosted relational trees. While our framework is limited to learning PCs with propositional logical constraints, it can be extended to use relational preference advice over variables. For example, the relational advice that ``\emph{Hypertensive disorders of pregnancy lead to pre-term birth}'' could be compiled to the propositional \emph{preference constraint} that subjects with the presence of any of chronic hypertension, gestational hypertension, and preeclampsia should have a higher conditional probability of pre-term birth. \citeauthor{Xu18SemanticLoss} (\citeyear{Xu18SemanticLoss}) use knowledge in the form of propositional logical formulas to learn deep neural networks using a semantic loss that computes the log probability of satisfying the formulas. Since this probability can be computed tractably in a PC, the semantic loss can be used to learn PCs using our framework.

The simple yet effective objective function we propose in Eq. \eqref{eq:opt4} has, in form, been successfully employed for tasks such as domain adaptation, domain generalization, and multiple class novelty detection \cite{sun2016deep, shi2022gradient, perera2019deep}.
We observed that augmenting the loss function with differentiable linear constraints is effective,
as shown in our experiments,
easy to be specified by domain experts, and allows the incorporation of a wide variety of constraints. While more complex objective functions are conceivable, our simple formulation encourages broader adoption.
Integration of more complex forms of domain knowledge, handling structured data (relational data), as well as learning the structure of a PC in a knowledge-intensive manner are promising future directions.

\section*{Acknowledgements}
The authors gratefully acknowledge the generous support by the NIH grant R01HD101246, the AFOSR award FA9550-23-1-0239, the ARO award W911NF2010224, and the DARPA Assured Neuro Symbolic Learning and Reasoning (ANSR)
award HR001122S0039.

\bibliography{aaai25-main}

\end{document}


\maketitle

%

\section{Encoding Privileged Information and Synergy}
We discuss two additional forms of knowledge and their corresponding constraints using the running example of modeling the risk of Gestational diabetes.
\begin{enumerate}
    \item Certain variables might not be available during test time. For example, measuring \textit{Genetic predisposition} to diabetes requires expensive testing and thus might not be available for patients at deployment time. However, if such features are
available at training time, they could be used to learn a more accurate distribution over the other features. We call constraints of this form \textbf{privileged information constraints.} These constraints can be represented as:
\begin{align*}
    \begin{aligned}
        P(x_i \mid \x_\text{obs - \{i\}} ) = P(x_i \mid  \x_{ \text{priv} \ \cup \ \text{obs - \{i\}}}),\quad \forall \x \in \mathcal{D} 
    \end{aligned}
\end{align*}
where $\X_\text{obs}$ denotes the set of variables observed during both training and deployment and $\X_\text{priv}$ denotes the set of variables observed only during training. 

Note that the privileged information constraint cannot be strictly satisfied. If it were satisfied exactly ($P(X_i \mid X_\text{obs})=P(X_i \mid X_\text{obs}\cup X_\text{priv})$) it would imply that the target ($X_i$) is independent of the privileged features ($X_\text{priv}$) given the observed features  ($X_\text{obs}$). This would only be possible if the privileged information were irrelevant. In the more realistic scenario where both observed and privileged information contribute to prediction, enforcing a soft constraint where $P(X_i \mid X_\text{obs})$ approximates $P(X_i \mid X_\text{obs}\cup X_\text{priv})$ can improve the accuracy of $P(X_i \mid X_\text{obs})$.

\item The combined effect of two variables can sometimes significantly enhance the likelihood of an outcome, more than their individual effects. 
    For example, the presence of high values of both \textit{Age} and \textit{BMI} can lead to a much higher risk of \textit{Gestational Diabetes} than high values of \textit{Age} or \textit{BMI} alone. We call such constraints \textbf{positive synergistic constraints}~\cite{YangEtAl2013}. They can be seen as a second-order monotonicity constraint.
    Specifically, the positive synergistic influence of $X_j$ and $X_k$ on $X_i$ (${X_j,X_k}_\prec^{S+} X_i$) can be represented as the linear inequality:
    \begin{align*}
        \begin{aligned}
            P(X_i \leq x'_i \mid x'_{j}, x_{k}) + P(X_i \leq x'_i \mid x_{j}, x'_{k}) > \\ 
            P(X_i \leq x_i \mid x_{j}, x_{k}) + P(X_i \leq x_i \mid x'_{j}, x'_{k}) \\\forall \x, \x' \text{ s.t. } x_i = x'_i,\ x'_j > x_j, x'_k > x_k
        \end{aligned}
    \end{align*}
\end{enumerate}

\begin{table*}[h]
    \centering
    \begin{tabular}{llp{0.6\linewidth}}
        \toprule
        \small
        \textbf{Name}                   & \textbf{Type} &  $\delta(\x, \x')$\\
        \midrule
        Generalization         & Equality & $P(\x) - P(\x')$ \\
        Privileged information & Equality & $P(x_i \mid \x_\text{obs - \{i\}} ) = P(x_i \mid  \x_{ \text{priv} \ \cup \ \text{obs - \{i\}}})$\\
        Context-specific independence & Equality & $P(x_i \mid x_j) - P(x_i \mid x_j, x_k)$ \\
        \midrule
        Class Imbalance (FN)          & Inequality & $P(X_i = 0 \mid \x_{-i}) - t_0  + \epsilon$ \\
        Monotonicity (${X_j}_{\prec}^{M+} X_i$)  & Inequality &$P(X_i \leq x_{i}' \mid x_{j}') - P(X_i \leq x_{i} \mid x_{j}) + \epsilon$\\
        Synergy (${X_j,X_k}_{\prec}^{S+} X_i$)   & Inequality & $P(X_i \leq x_i \mid x_{j}, x_{k}) + P(X_i \leq x_i \mid x'_{j}, x'_{k})$  $- P(X_i \leq x'_i \mid x'_{j}, x_{k}) - P(X_i \leq x'_i \mid x_{j}, x'_{k}) + \epsilon$ \\
        \bottomrule
    \end{tabular}
    \caption{Common forms of domain knowledge and their corresponding linear constraint representation in our framework.}
    \label{tab:fg}
\end{table*}

\begin{table*}[h]
    \centering
    \begin{tabular}{llp{\linewidth}}
        \toprule
        \small
        \textbf{Name} &  \textbf{Domain Set ($\mathcal{S}$)} \\
        \midrule
        Generalization & $\{ (\x, \x') \mid sim(\x,\x')\ \forall \x,\x'\in\mathcal{D}^2 \}$\\
        Privileged information & $\{ (\x, \x) \mid\ \forall \x\in\mathcal{D} \}$ \\
        Context-specific independence & $\{ (\x, \x) \mid \x_k = c\ \forall \x \in \text{Domain}(\X) \}$ \\
        Class imbalance (FN) & $\{ (\x, \x) \mid x_i = 1\ \forall \x \in\mathcal{D} \}$ \\
        Monotonicity (${X_j}_{\prec}^{M+} X_i$) & $\{ (\x, \x') \mid (x_{i}=x'_i) \land (x_{j}> x'_j) \land (\x_{-ij} = c) \ \forall \x, \x' \in \text{Domain}(\X)^2 \}$\\
        Synergy (${X_j,X_k}_{\prec}^{S+} X_i$)  & $\{ (\x, \x') \mid (x_{i}=x'_i) \land (x_{j}> x'_j) \land (x_{k}> x'_k) \land (\x_{-ijk} = c) \ \forall \x, \x' \in \text{Domain}(\X)^2 \}$ \ \ \ \ \ \ \ \ \\
        \bottomrule
    \end{tabular}
    \caption{Domain sets ($\mathcal{S}$) over which the penalty functions encoding common forms of domain knowledge are evaluated.}
    \label{tab:selectors}
\end{table*}

\section{Details on Experimental Setup}
We implemented both \textit{RatSPN} and \textit{EinsumNet} in pytorch, adapting from the implementation of \cite{peharz_20_einsum,peharz20a-rat-spn}. We used gradient descent with an Adam optimizer to train all models. 

\subsection{Model Architecture and Hyperparameters}
To maintain consistency and fairness in our comparative analysis, we adopted identical architectures and model capacity-determining hyperparameters for models trained both with and without domain constraints. These hyperparameters, which define the structural complexity and learning potential of \textit{RatSPN} and \textit{EinsumNet}, include:
\begin{itemize}[wide,labelwidth=!, labelindent=0pt]
    \item \textit{Circuit Depth} $(D)$: The number of layers in the probabilistic circuit, influencing its ability to capture complex interactions.
    \item \textit{Input Leaf Distributions per Variable} $(I)$: The number of distinct input leaf distributions assigned to each variable.
    \item \textit{Sum Units per Sum Node} $(S)$: Determines the number of vectorized distributions at each sum node.
    \item \textit{Number of Replicas} $(R)$: The number of repeated ensemble structures within the probabilistic circuit, it enhances the model's representational power.
    \item \textit{Leaf Distribution Type}: The type of simple and tractable probability distribution used at the leaves of the circuit.
\end{itemize}

For an in-depth explanation of these hyperparameters and their roles in the models, please refer to the original works \cite{peharz20a-rat-spn,peharz_20_einsum}. In addition to the above structural hyperparameters, training-specific hyperparameters include learning rate $(lr)$ used by the optimizer, batch size $(B)$, and the total number of training epochs $(E)$. For the experiments in which we incorporate domain constraints, we have the following adiitional hyperparameters:
\begin{itemize}[wide,labelwidth=!, labelindent=0pt]
    \item Penalty Weight $(\lambda)$: The factor that determines the trade-off between fitting the data and satisfying the domain constraints.
    \item Penalty Weight Control Parameter $(\gamma)$: This parameter aids in transitioning between soft and hard constraints. If $\gamma$ is set, then $\lambda$ is iteratively increased to make the constraints satisfied to the maximum extent.
\end{itemize}

For the experiments, we utilized Intel Xeon Platinum 8167M CPU with 24 cores along with 2 NVIDIA Tesla V100 GPUs, each with 16GB memory. The datasets chosen for our experiments cover a wide array, each selected to challenge and validate the efficacy of domain constraints in different contexts. In the subsequent subsections, we provide comprehensive details on these datasets, including their characteristics and the specific hyperparameters employed for each. 
 
\subsection{Tabular data sets}

Table \ref{tab:data} shows the number of variables, the number of examples, and the number of constraints for data sets derived from Bayesian Networks (BN), UCI benchmark data sets (UCI) and the Real-world clinical data sets (RW). Due to the small size of the tabular data sets, we used the hyperparameters D = 2, I = 20, S = 20, and R = 5. Additionally, we set $lr=0.001, E=100, \gamma = 10$ , and the iteration threshold $t_\text{max} = 10$ for all the tabular data sets. We used a batch size (B) of 100 for the BN data sets and 64 for all of the other tabular data sets.

Rows 1--4 show the details of the data sets derived from BNs. Figure \ref{fig:bn-viz} provides a visualization of the four BNs we used - \textit{asia}, \textit{sachs}, \textit{survey}, and \textit{earthquake} BNs, for validating the effectiveness of our framework in faithfully integrating domain knowledge as constraints into the learning of PCs. The BNs were taken from the widely used \textit{bnlearn} repository (\url{https://www.bnlearn.com/bnrepository/}). We sampled $200$ data point from each BN, and used half as the training set and the remaining half as the test set. We encoded domain knowledge by taking 2 Conditional Independence relations from each BN. Each Conditional Independence relation of the form $X \perp\!\!\!\!\perp Y \mid Z$ is equivalent to the set of Context-specific independence relations (CSIs) $$\{ (X \perp\!\!\!\!\perp Y \mid Z=z) : \forall z \in \text{Domain}(Z) \}$$
We encode each CSI of the form $(X \perp\!\!\!\!\perp Y \mid Z=z)$ as the constraints
\begin{align}
\begin{aligned}
    P(X=x \mid Y=y, Z=z) = P(X=x \mid Z=z) \\
    P(Y=y \mid X=x, Z=z) = P(Y=y \mid Z=z) \\
    \forall x \in \text{Domain}(X), y \in \text{Domain}(Y)
\end{aligned}
\end{align}
Table \ref{tab:knowledge} lists the domain knowledge corresponding to each of the data sets used for experiments.

\tikzset{
latent/.style={
draw, circle,
minimum size=1cm,
node distance=1cm,}}
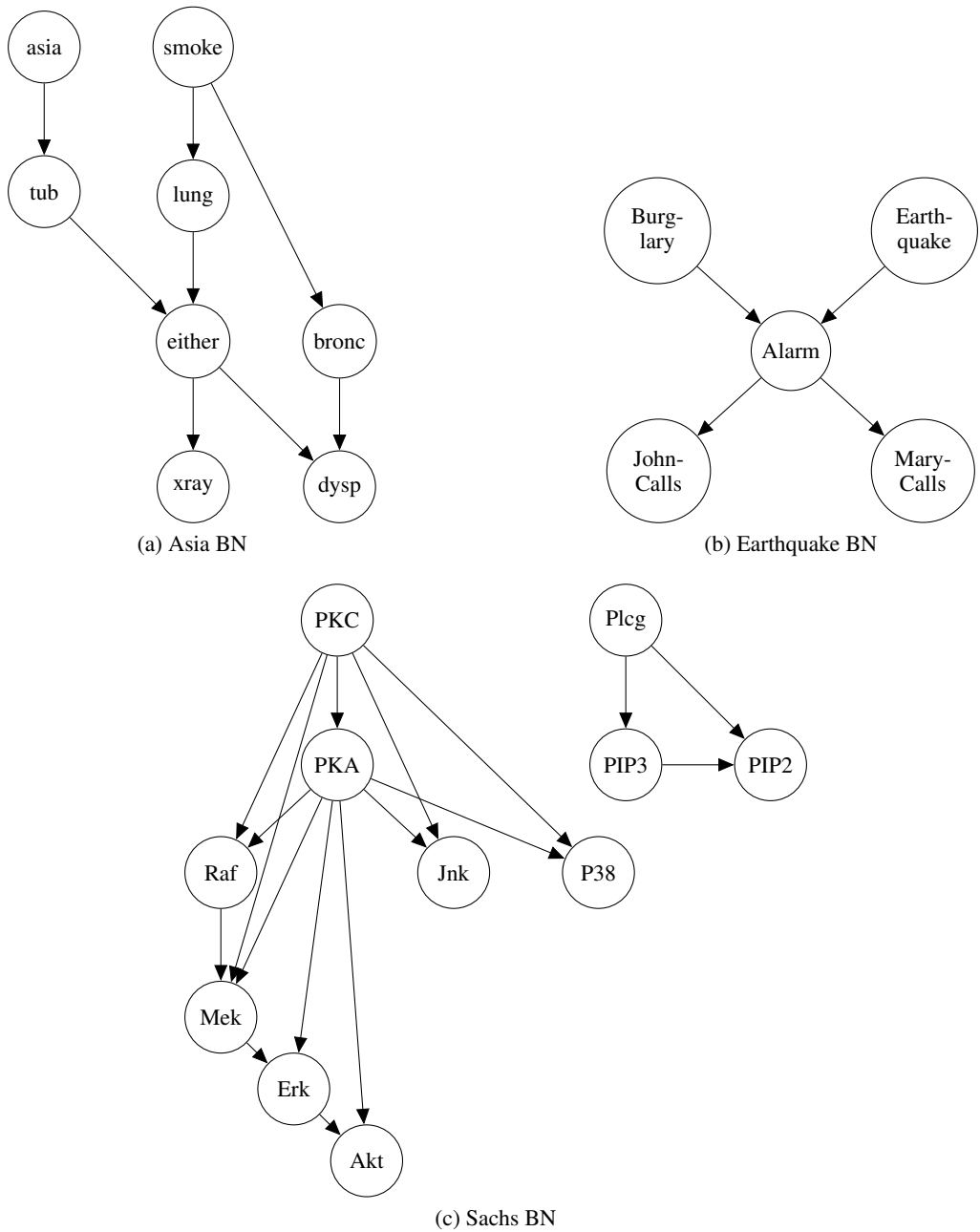
\begin{figure*}
    \centering
    \small
    \begin{tabular}{ccc}
    \subfloat[Asia BN]{\begin{tikzpicture}
      \node[latent] (asia) {asia};
      \node[latent, below=of asia] (tub) {tub};
      \node[latent, right=of asia] (smoke) {smoke};
      \node[latent, below=of smoke] (lung) {lung};
      \node[latent, below=of lung] (either) {either};
      \node[latent, below=of lung,right=of either] (bronc) {bronc};
      \node[latent, below=of either] (xray) {xray};
      \node[latent, below=of bronc] (dysp) {dysp};
    
      \edge {asia} {tub}
      \edge {tub} {either}
      \edge {smoke} {lung}
      \edge {smoke} {bronc}
      \edge {lung} {either}
      \edge {either} {xray,dysp}
      \edge {bronc} {dysp}
    \end{tikzpicture}} \hspace{3cm}
    \subfloat[Earthquake BN]{
    \begin{tikzpicture}
      \node (n1){};
      \node[latent, left=of n1, text width=1cm, align=center] (burglary) {Burg-\\lary};
      \node[latent, right=of n1,text width=1cm,align=center] (earthquake) {Earth-\\quake};
      \node[latent, below=of n1] (alarm) {Alarm};
       \node[below=of alarm](n2){};
       \node[latent, left=of n2,text width=1cm,align=center] (johncalls) {John-\\Calls};
       \node[latent, right=of n2,text width=1cm,align=center] (marycalls) {Mary-\\Calls};
      
      \edge {burglary} {alarm}
      \edge {earthquake} {alarm}
      \edge {alarm} {johncalls}
      \edge {alarm} {marycalls}
    \end{tikzpicture}
    }\\
    \subfloat[Sachs BN]{
    \begin{tikzpicture}
      \node[latent](pkc){PKC};
      
      \node[latent, right=of pkc, xshift=2cm] (plcg) {Plcg};
      
      \node[latent, below=of plcg] (pip3) {PIP3};
      \node[latent, right=of pip3] (pip2) {PIP2};
      
      \node[latent, below=of pkc] (pka) {PKA};
      \node[below of=pka, yshift=-0.5cm](n1) {};
      \node[latent, left=of n1] (raf) {Raf};
      \node[latent, right=of n1] (jnk) {Jnk};
      \node[latent, right=of jnk] (p38) {P38};

      \node[latent, below=of raf] (mek) {Mek};
       \node[latent, right=of mek, xshift=-1cm, yshift=-1cm] (erk) {Erk};
       \node[latent, right=of erk, xshift=-1cm, yshift=-1cm] (akt) {Akt};

      \edge {plcg} {pip3}
      \edge {plcg} {pip2}
      \edge {pip3} {pip2}

      \edge {pkc} {pka}
      \edge {pkc} {raf}
      \edge {pkc} {jnk}
      \edge {pkc} {p38}
      \edge {pkc} {mek}
      \edge {pka} {raf}
      \edge {pka} {mek}
      \edge {pka} {erk}
      \edge {pka} {akt}
      \edge {pka} {jnk}
      \edge {pka} {p38}
      \edge {raf} {mek}
      \edge {mek} {erk}
      \edge {erk} {akt}
    \end{tikzpicture}
    }
    \end{tabular}
    \subfloat[Survey BN]{
    \begin{tikzpicture}
      \node[](n1) {};
      \node[latent, left= of n1,text width=1cm, align=center](a){Age};
      \node[latent, right = of n1,text width=1cm, align=center](s){Sex};
      \node[latent, below of = n1,text width=1cm, align=center](e){Educ-\\ation};
      \node[below = of e](n2) {};
      \node[latent, left=of n2,text width=1cm, align=center](o){Occu-\\pation};
      \node[latent, right=of n2,text width=1cm, align=center](r){Resi-\\dence};
      \node[latent, below = of n2,text width=1cm, align=center](t){Tra-\\vel};

      \edge {a} {e}
      \edge {s} {e}
      \edge {e} {o}
      \edge {e} {r}
      \edge {o} {t}
      \edge {r} {t}
     \end{tikzpicture}
    }
    \caption{Visualization of the $4$ \textbf{Bayesian Networks} (BN) used for generating data to validate the effectiveness of our approach in integrating domain knowledge in the form of constraints. For more information regarding each BN, please refer to the \textit{bnlearn} repository (\url{https://www.bnlearn.com/bnrepository/}).}
    \label{fig:bn-viz}
\end{figure*}

Rows 4--7 of table \ref{tab:data} show the details of the UCI data sets and rows 8-9 show the details of the Real-world clinical data sets. We split each of these data sets into equally sized training and testing sets. We used conditional independences and monotonic influence statements from prior work\cite{sup-YangEtAl2013,sup-Mathur2023KICN,Mathur2024AIME} as constraints.

\begin{table}[ht!]
    \centering
    \begin{tabular}{llrrrr}
    \toprule
    & Name & $\|X\|$ & $\|D\|$ & CIs & MIS\\
    \midrule
    BN & asia & 8 & 200 & 2 & 1 \\
    & sachs & 11 & 200 &  2 & 0 \\
    & earthquake & 5 & 200 & 2 & 1 \\
    & survey & 6 & 200 & 2 & 1 \\
    \midrule
    UCI & breast-cancer & 10 & 277 & 0 & 5 \\
    & diabetes & 9 & 392 & 0&  4 \\
    & thyroid & 6 & 185 & 0 & 5 \\
    & heart-disease & 6 & 297 &0 & 5 \\
    \midrule
    RW & numom2b-a & 8 & 3657 & 0 & 6 \\
     & numom2b-b & 7 & 9368 & 1 & 1 \\
    \bottomrule
    \end{tabular}
    \caption{The number of variables ($\|X\|$), the number of examples ($\|D\|$) and the number of constraints (CIs and MISs) for each data set  }
    \label{tab:data}
\end{table}

\begin{table*}[ht!]
    \centering
    \begin{tabular}{llp{14cm}}
    \toprule
    & Name & Knowledge \\
    \midrule
    BN & asia & \{  asia $\perp\!\!\!\!\perp$ bronc $\mid$ lung, asia $\perp\!\!\!\!\perp$ bronc $\mid$ smoke, either$_\prec^{M+}$xay \} \\
    & sachs & \{ Akt $\perp\!\!\!\!\perp$ PIP2 $\mid$ Erk, Akt $\perp\!\!\!\!\perp$ PIP2 $\mid$ Jnk \}\\
    & earthquake & \{ Burglary $\perp\!\!\!\!\perp$ JohnCalls $\mid$ Alarm, Burglary $\perp\!\!\!\!\perp$ MaryCalls $\mid$ Alarm, Alarm$_\prec^{M+}$MaryCalls \}\\
    & survey & \{ Age $\perp\!\!\!\!\perp$ Occupation $\mid$ Education, Age $\perp\!\!\!\!\perp$ Residence $\mid$ Education, Education$_\prec^{M+}$Residence \} \\
    \midrule
    UCI & breast-cancer & \{ age$_\prec^{M+}$recurrence, menopause$_\prec^{M+}$recurrence, deg\_malig$_\prec^{M+}$recurrence, tumor\_size$_\prec^{M+}$recurrence, irradiat$_\prec^{M-}$recurrence \} \\
    & diabetes & \{ Age$_\prec^{M+}$Diabetes, Pregnancies$_\prec^{M+}$Diabetes, BMI$_\prec^{M+}$Diabetes, PedigreeFunction$_\prec^{M+}$Diabetes \}  \\
    & thyroid & \{T3\_resin$_\prec^{M+}$HyperThyroid, T3$_\prec^{M+}$HyperThyroid, TSH$_\prec^{M+}$HyperThyroid, TSH\_diff$_\prec^{M+}$HyperThyroid, T4$_\prec^{M+}$HyperThyroid\}   \\
    & heart-disease & \{sex\_male$_\prec^{M+}$HD, age$_\prec^{M+}$HD, trestbps$_\prec^{M+}$HD, chol$_\prec^{M+}$HD, diabetes$_\prec^{M+}$HD\}  \\
    \midrule
    RW & numom2b-a & $\{\text{BMI}_{\prec}^{M+}\text{GD}$, $\text{METs}_{\prec}^{M-}\text{GD}$, $\text{Age}_{\prec}^{M+}\text{GD},$ $\text{Hist}_{\prec}^{M+}\text{GD},$ $\text{PCOS}_{\prec}^{M+}\text{GD},$ $\text{HiBP}_{\prec}^{M+}\text{GD} \}$\\
     & numom2b-b & \{ Age $\perp\!\!\!\!\perp$ BMI $\mid$ HiBP, PreEc$_\prec^{M+}$PTB \} \\
    \bottomrule
    \end{tabular}
    \caption{The domain knowledge corresponding to each of the data sets }
    \label{tab:knowledge}
\end{table*}

\subsection{Helix}
We adapted the synthetic 3D Helix dataset \cite{vqflows-sidheekh22a} comprising of 10000 train, 5000 validation and 5000 test datapoints. Each datapoint is of the form $\{x,y,z\}$ representing the position of the datapoint along x-axis, y-axis and z-axis, respectively. The training data is generated by sampling 10000 points uniformly over the range $(0,2\pi)$ to represent $x$. Similarly, validation and test datasets are generated by sampling 5000 points uniformly over the range $(0, 4\pi)$ to represent $x$. The coordinates for $y$ and $z$ for each corresponding $x$ are computed as follows:
\begin{align*}
    y \sim \mathcal{N}(sin(x), 1) \\
    z \sim \mathcal{N}(cos(x), 1) \\
\end{align*}
This leads to the validation and test datasets having datapoints in $x$ range of $(2\pi, 4\pi)$ that are not present in the training data. This allows us to compare the effects of naively adding a few datapoints from the unseen region to the training data versus leveraging these additional datapoints to encode domain knowledge.
The intricate spiral structure of this 3D manifold presents a challenging learning scenario while allowing us to demonstrate the effectiveness of domain knowledge provided in the form for a Generalization Constraint (GC) to encode the symmetrical nature of the manifold. A 3D helix consists of a structural motif - a spiral of length $2\pi$ along the x-axis - that repeats itself. So, it is natural to expect points that are $2\pi$ units apart along the x-axis and lie on a spiral structure to have equal (or nearly equal) log-likelihoods w.r.t the model. This domain knowledge is easily encoded as a GC. 

\subsection{Set Representation Datasets} 

\begin{figure*}[ht]
    \centering
    \begin{tabular}{cc}
    \subfloat[Set-MNIST-Full]{ \includegraphics[width=0.23\linewidth]{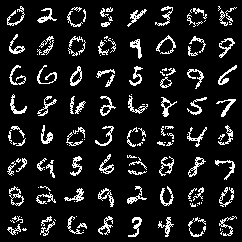}}\hspace{2.5pt}
    \subfloat[Set-MNIST-Odd]{\includegraphics[width=0.23\linewidth]{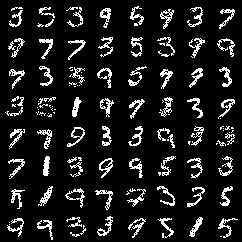}}\hspace{2.5pt}
    \subfloat[Set-MNIST-Even]{ \includegraphics[width=0.23\linewidth]{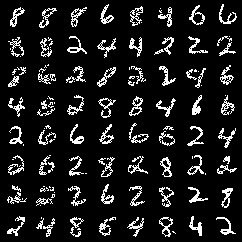}}\hspace{2.5pt}
    \subfloat[Set-Fashion-MNIST]{\includegraphics[width=0.23\linewidth]{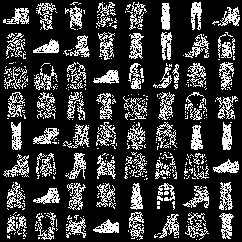}}
    \end{tabular}
    \caption{ \textbf{Datasets}: Visualization of randomly sampled datapoints from the (a) Set-MNIST-Full, (b) Set-MNIST-Odd, (c) Set-MNIST-Even, and (d) Set-Fashion-MNIST datasets.}
    \label{fig:set-mnist-datasets}
\end{figure*}

\textbf{Set-MNIST.}
To adapt the classic MNIST dataset \cite{mnist} of $28\times28 = 784$ pixel handwritten digits for our experiments with set-based representations, we transformed the original images into 2D point clouds. This was achieved by sampling the coordinates of $100$ foreground pixels (i.e., non-zero pixels) from each image, following the methodology outlined in \cite{zhang2019deep}. Consequently, each of the $100$ features in this transformed dataset represents a discrete variable with possible values ranging between $0$ and $784$, accounting for the one-dimensional unfolding of the 2D pixel grid. Images with fewer than $100$ foreground pixels were excluded from the dataset to maintain uniformity. This process gave rise to what we refer to as the Set-MNIST-Full dataset, which had $31,061$ training data points and $5,360$ test data points.
To explore the effects of data scarcity on model performance and the potential benefits of incorporating domain knowledge, we divided Set-MNIST-Full into the following two subsets: (1) Set-MNIST-Even: Containing only set representations of even digits and comprising $11,409$ training data points and $1,964$ test data points.
(2) Set-MNIST-Odd: Containing only set representations of odd digits and comprising $14,453$ training data points and $2,529$ test data points.

\textbf{Set-Fashion-MNIST.} 
In a similar fashion as above, we processed the Fashion-MNIST \cite{fashion-mnist} dataset into a set-based format, termed Set-Fashion-MNIST. This involved sampling locations of $200$ foreground pixels from each image to form a point cloud representation, yielding a dataset with $35,639$ training data points and $5,940$ test data points, each described by $200$ dimensions. A validation set was created for each set dataset from random permutations of the train split to facilitate model tuning and evaluation. Figure \ref{fig:set-mnist-datasets} provides visualizations of the transformed Set-MNIST and Set-Fashion-MNIST datasets.

For our experiments on the Set-MNIST and Set-Fashion-MNIST datasets, we chose to implement a categorical distribution with 784 categories as the leaf distribution for both the \textit{EinsumNet} and \textit{RatSPN} models. The architectural hyperparameters were uniformly set as $D=6$, $I=10$, $S=10$ and $R=10$, to ensure a complex model capacity capable of capturing the nuances of the set-based data. The learning rate ($lr$) was set at 0.1, with a batch size ($B$) of 100 and a total of 200 epochs ($E$) for the training process.

To effectively integrate the concept of permutation invariance into our models as a generalization constraint, we augmented the domain set by including 2 permutations for each data point. The penalty weight ($\lambda$) and the penalty weight control parameter ($\gamma$) were both set to 1, striking a balance between soft and hard constraint application and ensuring that the constraints significantly influenced the learning process without overshadowing the primary data-driven objectives. Considering the vast number of possible permutations for each data point, which is $100!$ or approximately $9.332622 \times 10^{157}$, the inclusion of just 2 permutations per data point in the domain set proved to be sufficient for a notable improvement in generalization performance. This demonstrates the efficacy of our approach in leveraging a small, representative subset of the domain set to enforce the permutation invariance constraint effectively.

\section{Additional Results}
\begin{figure*}
    \centering
    \begin{tikzpicture} 
    \node[rectangle,draw=white!90,fill=white,opacity=1,minimum width=0.5cm,minimum height=0.65cm] at (0,0) {};
    \node[rectangle,draw=white!90,fill=gray!10,opacity=0.9,minimum width=3.cm,minimum height=0.65cm] at (2.05,0) {\textit{RatSPN}};
    \node[rectangle,draw=white!90,fill=gray!10,opacity=0.9,minimum width=3.cm,minimum height=0.65cm] at (7.45,0) {\textit{EinsumNet}};
    \end{tikzpicture}
    
    \includegraphics[width=0.33\linewidth]{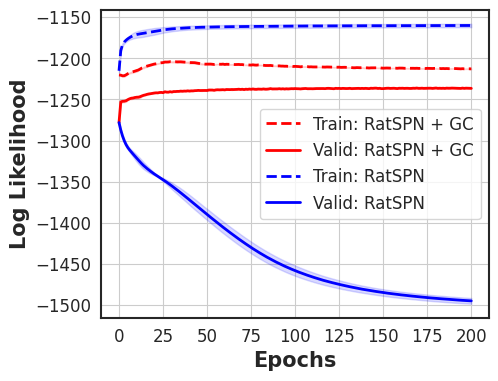}
    \includegraphics[width=0.33\linewidth]{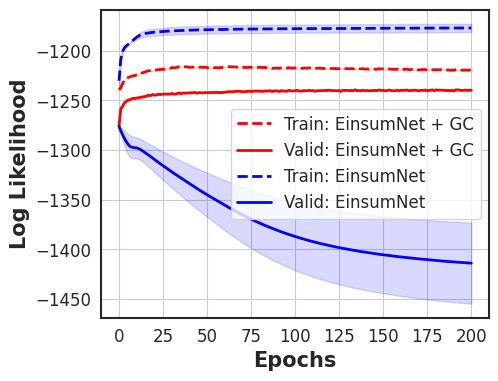}
    \caption{\textbf{Learning Curves:} Mean train and validation log-likelihoods of \textit{RatSPN}  and \textit{EinsumNet} trained \textcolor{red}{with} (in red) and \textcolor{blue}{without} (in blue) incorporating Generalization Constriant (GC) on the \textbf{Set-Fashion-MNIST} dataset, across training epochs.  \textit{EinsumNet} and \textit{RatSPN} overfits to the training data and hence is unable to generalize on the validation set, whereas incorporating permutation invariance as a Generalization Constraint (GC) helps the models capture symmetries to achieve better generalization performance. The shaded region denotes the standard deviation across $3$ trials.}
    \label{fig:set-fmnist-learning-curves}
\end{figure*}

\begin{figure*}
    \centering
    \begin{tikzpicture} 
    \node[rectangle,draw=white!90,fill=white,opacity=1,minimum width=0.3cm,minimum height=0.65cm] at (0,0) {};
    \node[rectangle,draw=white!90,fill=gray!10,opacity=0.9,minimum width=3.cm,minimum height=0.65cm] at (1.0,0) {Set-MNIST-Even};
    \node[rectangle,draw=white!90,fill=gray!10,opacity=0.9,minimum width=3.cm,minimum height=0.65cm] at (6.75,0) {Set-MNIST-Odd};
    \node[rectangle,draw=white!90,fill=gray!10,opacity=0.9,minimum width=3.cm,minimum height=0.65cm] at (12.9,0) {Set-MNIST-Full};
    \end{tikzpicture}
    \includegraphics[width=0.32\linewidth]{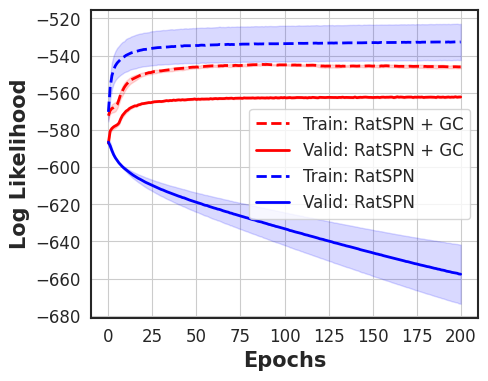}
    \includegraphics[width=0.32\linewidth]{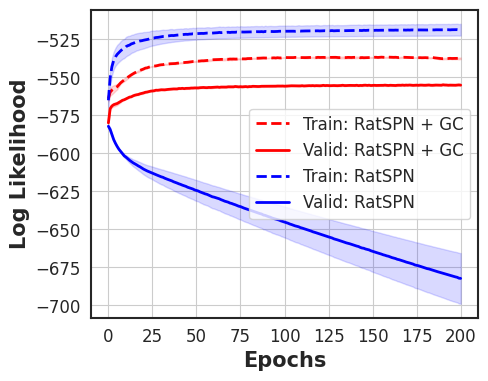}
    \includegraphics[width=0.32\linewidth]{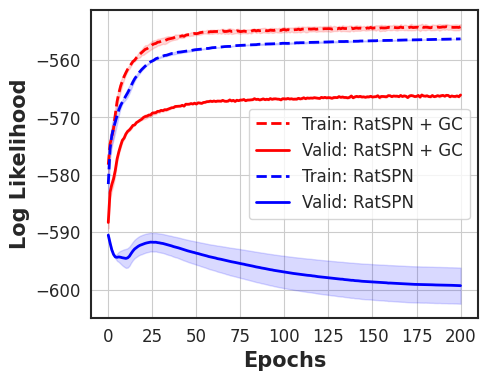}
    
    \includegraphics[width=0.32\linewidth]{figures/set-mnist/set-mnist-even_EinsumNet_learning_curve.png}
    \includegraphics[width=0.32\linewidth]{figures/set-mnist/set-mnist-odd_EinsumNet_learning_curve.png}
    \includegraphics[width=0.32\linewidth]{figures/set-mnist/set-mnist-full_EinsumNet_learning_curve.png}
    \caption{\textbf{Learning Curves:} Mean train and validation log-likelihoods of \textit{RatSPN} (top row) and \textit{EinsumNet} (bottom row)  trained \textcolor{red}{with} (in red) and \textcolor{blue}{without} (in blue) incorporating Generalization Constriant (GC) on the three \textbf{Set-MNIST} datasets, across training epochs.  \textit{EinsumNet} and \textit{RatSPN} overfits to the training data and hence is unable to generalize on the validation set, whereas incorporating permutation invariance as a Generalization Constraint (GC) helps the models capture symmetries to achieve better generalization performance. The shaded region denotes the standard deviation across $3$ independent trials.}
    \label{fig:set-mnist-learning-curves}
\end{figure*}

\begin{figure*}[b!]
    \centering
    \begin{tikzpicture} 
    \node[rectangle,draw=white!90,fill=white,opacity=1,minimum width=0.5cm,minimum height=0.5cm] at (0,0) {};
    \node[rectangle,draw=white!90,fill=gray!10,opacity=0.9,minimum width=3.5cm,minimum height=0.5cm] at (2.35,0) {Set-MNIST-Even};
    \node[rectangle,draw=white!90,fill=gray!10,opacity=0.9,minimum width=3.5cm,minimum height=0.5cm] at (6.6,0) {Set-MNIST-Odd};
    \node[rectangle,draw=white!90,fill=gray!10,opacity=0.9,minimum width=3.5cm,minimum height=0.5cm] at (10.8,0) {Set-MNIST-Full};
    \end{tikzpicture}
    
    \begin{tikzpicture}
        \node [draw=black!2,rotate=90,anchor=center,fill=gray!8] { \hspace{0.085\linewidth} Real Data\hspace{0.085\linewidth} };    
    \end{tikzpicture}
    \includegraphics[width=0.3\linewidth]{figures/set-mnist/real_data/set-mnist-even.png}
    \includegraphics[width=0.3\linewidth]{figures/set-mnist/real_data/set-mnist-odd.png}
    \includegraphics[width=0.3\linewidth]{figures/set-mnist/real_data/set-mnist-full.png}
    
    \begin{tikzpicture}
        \node [draw=black!2,rotate=90,anchor=center,fill=gray!8] { \hspace{0.075\linewidth} EinsumNet\hspace{0.078\linewidth} };    
    \end{tikzpicture}
    \includegraphics[width=0.3\linewidth]{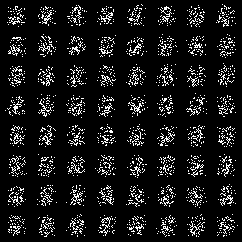}
    \includegraphics[width=0.3\linewidth]{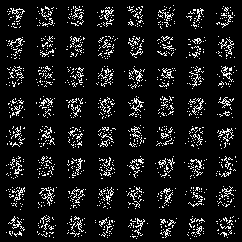}
    \includegraphics[width=0.3\linewidth]{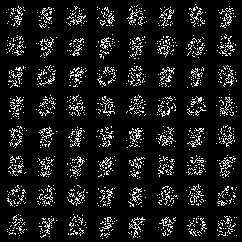}

    \begin{tikzpicture}
        \node [draw=black!2,rotate=90,anchor=center,fill=gray!8] { \hspace{0.05\linewidth} EinsumNet + GC\hspace{0.05\linewidth} };    
    \end{tikzpicture}
    \includegraphics[width=0.3\linewidth]{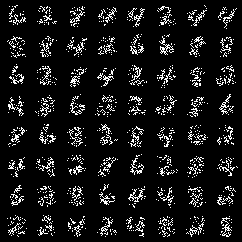}
    \includegraphics[width=0.3\linewidth]{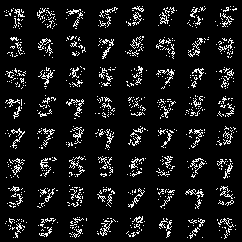}
    \includegraphics[width=0.3\linewidth]{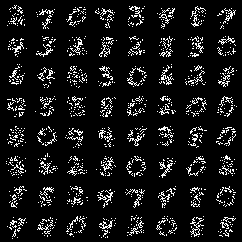}
    
    \caption{ Visualization of datapoints generated by \textit{EinsumNet} trained on the three Set-MNIST datasets with (bottom row) and without (middle row) the generalization constraint. The top row visualizes real data samples drawn from the corresponding datasets. The incorporation of Permutation Invariance as a Generalization Constraint helps the model generate higher fidelity samples better resembling the real data distribution.}
    \label{fig:set-mnist-einsumnet-generated}
\end{figure*}

\begin{figure*}[t]
    \centering
    \begin{tikzpicture} 
    \node[rectangle,draw=white!90,fill=white,opacity=1,minimum width=0.5cm,minimum height=0.5cm] at (0,0) {};
    \node[rectangle,draw=white!90,fill=gray!10,opacity=0.9,minimum width=3.5cm,minimum height=0.5cm] at (2.35,0) {Set-MNIST-Even};
    \node[rectangle,draw=white!90,fill=gray!10,opacity=0.9,minimum width=3.5cm,minimum height=0.5cm] at (6.6,0) {Set-MNIST-Odd};
    \node[rectangle,draw=white!90,fill=gray!10,opacity=0.9,minimum width=3.5cm,minimum height=0.5cm] at (10.8,0) {Set-MNIST-Full};
    \end{tikzpicture}
    
    \begin{tikzpicture}
        \node [draw=black!2,rotate=90,anchor=center,fill=gray!8] { \hspace{0.08\linewidth} Real Data\hspace{0.08\linewidth} };    
    \end{tikzpicture}
    \includegraphics[width=0.3\linewidth]{figures/set-mnist/real_data/set-mnist-even.png}
    \includegraphics[width=0.3\linewidth]{figures/set-mnist/real_data/set-mnist-odd.png}
    \includegraphics[width=0.3\linewidth]{figures/set-mnist/real_data/set-mnist-full.png}
    
    \begin{tikzpicture}
        \node [draw=black!2,rotate=90,anchor=center,fill=gray!8] { \hspace{0.09\linewidth} RatSPN \hspace{0.09\linewidth} };    
    \end{tikzpicture}
    \includegraphics[width=0.3\linewidth]{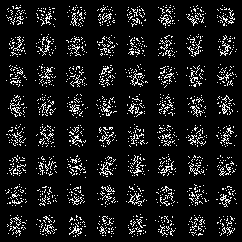}
    \includegraphics[width=0.3\linewidth]{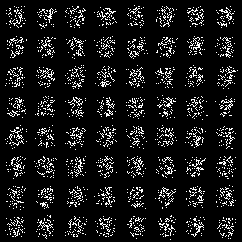}
    \includegraphics[width=0.3\linewidth]{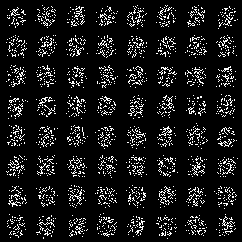}

    \begin{tikzpicture}
        \node [draw=black!2,rotate=90,anchor=center,fill=gray!8] { \hspace{0.06\linewidth} RatSPN + GC\hspace{0.06\linewidth} };    
    \end{tikzpicture}
    \includegraphics[width=0.3\linewidth]{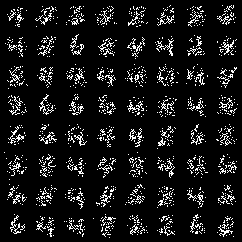}
    \includegraphics[width=0.3\linewidth]{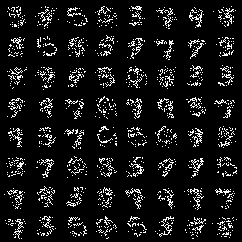}
    \includegraphics[width=0.3\linewidth]{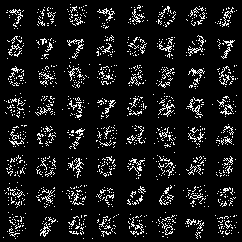}
    
    \caption{ Visualization of datapoints generated by \textit{RatSPN} trained on the three Set-MNIST datasets with (bottom row) and without (middle row) the generalization constraint. The top row visualizes real data samples drawn from the corresponding datasets. The incorporation of Permutation Invariance as a Generalization Constraint helps the model generate higher fidelity samples better resembling the real data distribution.}
    \label{fig:set-mnist-ratspn-generated}
\end{figure*}

We provide visualizations of the learning curves of \textit{EinsumNet} and \textit{RatSPN} trained with and without the Generalization Constraint (GC) on the Set-MNIST datasets in Figure \ref{fig:set-mnist-learning-curves} and on the Set-Fashion-MNIST dataset in Figure \ref{fig:set-fmnist-learning-curves}. We can observe that in the absence of the domain knowledge, all the models without $GC$ overfit to the training data, as the train log-likelihood increases while the validation log-likelihood decreases,  while incorporation of $GC$ helps them generalize better. Qualitative Visualizations of the samples generated by each of these models is provided in Figures \ref{fig:set-mnist-einsumnet-generated} - \ref{fig:set-mnist-ratspn-generated}. Clearly, models with $GC$ incorporated are able to learn the real distribution better to generate higher fidelity samples.

\subsubsection{Ablation Study on Synthetic Dataset.} 
\begin{figure*}[ht]
    \centering
    \begin{tabular}{ccc}
    \subfloat[50]{ \includegraphics[width=0.18\linewidth]{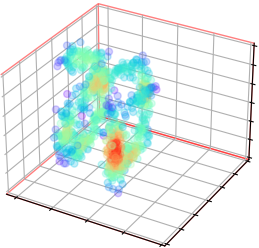}}\hspace{2.5pt}
    \subfloat[100]{\includegraphics[width=0.18\linewidth]{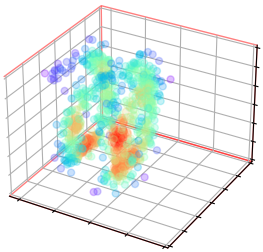}}\hspace{2.5pt}
    \subfloat[150]{\includegraphics[width=0.18\linewidth]{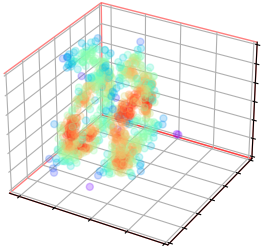}}\hspace{2.5pt}
    \subfloat[200]{\includegraphics[width=0.18\linewidth]{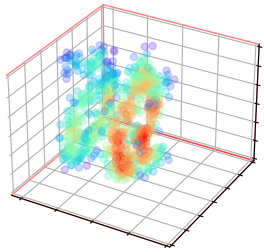}}\hspace{2.5pt}
    \subfloat[250]{\includegraphics[width=0.18\linewidth]{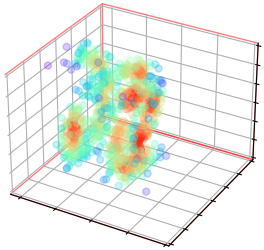}}\hspace{2.5pt} \\
    \subfloat[300]{ \includegraphics[width=0.18\linewidth]{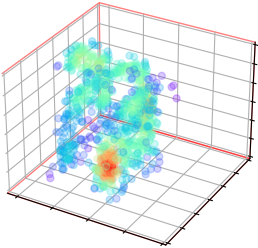}}\hspace{2.5pt}
    \subfloat[350]{\includegraphics[width=0.18\linewidth]{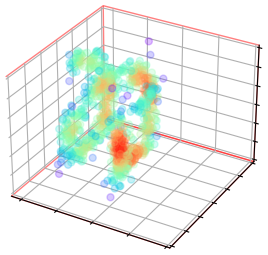}}\hspace{2.5pt}
    \subfloat[400]{\includegraphics[width=0.18\linewidth]{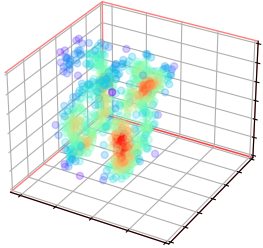}}\hspace{2.5pt}
    \subfloat[450]{\includegraphics[width=0.18\linewidth]{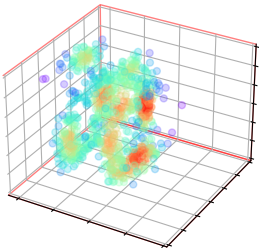}}\hspace{2.5pt}
    \subfloat[500]{\includegraphics[width=0.18\linewidth]{figures/ablation/helix/constrained_helix_ablation_50.png}}\hspace{2.5pt}
    \end{tabular}
    \caption{\small \textbf{Domain set size ablation on 3D Helix dataset}: Visualization of samples generated on EinsumNet+GC on the 3D Helix dataset. The labels denote the domain set size.}
    \label{fig:3D_helix_ablation}
\end{figure*}

We also performed an additional ablation to study the impact of domain set size on learning with and without GC on the 3D Helix dataset. We varied the domain set size in the range $(50, 500)$ in increments of $50$. Figure \ref{fig:3D_helix_ablation} shows the samples generated by \textit{EinsumNet+GC} on the 3D Helix dataset. The generated samples are similar to the test dataset for all domain set sizes ranging from 50 to 500, implying model performance is not particularly sensitive to the domain set size. This suggests that few high quality samples used to encode domain knowledge are sufficient for the model to generalize to unseen test regions.

\subsubsection{Penalty violation for baseline.}
\begin{table}[ht]
\centering
\begin{tabular}{ll}
\toprule
Dataset Name       & Penalty Violation Score - $\zeta(\mathcal{M})$ \\
\midrule
asia       & $0.02772$ \\
earthquake & $0.03373$ \\
sachs      & $0.02671$ \\
survey     & $0.02666$ \\ 
\bottomrule
\end{tabular}
\caption{Penalty violation score for baseline RAT-SPN model learned from the BN datasets}\label{tab:baseline-penalty-violation}
\end{table}

To add further context to the empirical results for (Q1), we present the degree of penalty violations for the baseline RAT-SPN model learned from BN data sets without domain knowledge in table \ref{tab:baseline-penalty-violation}.


\bibliography{aaai25-supp}